\documentclass[preprint,10pt,5p,twocolumn,authoryear]{elsarticle}

\usepackage{amssymb}
\usepackage{amsmath}
\usepackage{amssymb}
\usepackage{amsmath}
\usepackage{caption} %改变图表标题
\usepackage{booktabs} %调整表格线与上下内容的间隔
\usepackage{longtable}%调用跨页表格
\usepackage{multirow} %多行合并
\usepackage{array} %调用公式宏包的命令应放在调用定理宏包命令之前，也能控制表格
\usepackage{graphicx}
\usepackage{subcaption}
\usepackage{float}
\usepackage{hyperref}
\usepackage{multicol}
\usepackage{lipsum} % 用于生成示例文本
\journal{Medical Image Analysis}

\begin{document}

\begin{frontmatter}

%% Title, authors and addresses

%% use the tnoteref command within \title for footnotes;
%% use the tnotetext command for theassociated footnote;
%% use the fnref command within \author or \affiliation for footnotes;
%% use the fntext command for theassociated footnote;
%% use the corref command within \author for corresponding author footnotes;
%% use the cortext command for theassociated footnote;
%% use the ead command for the email address,
%% and the form \ead[url] for the home page:
%% \title{Title\tnoteref{label1}}
%% \tnotetext[label1]{}
%% \author{Name\corref{cor1}\fnref{label2}}
%% \ead{email address}
%% \ead[url]{home page}
%% \fntext[label2]{}
%% \cortext[cor1]{}
%% \affiliation{organization={},
%%            addressline={}, 
%%            city={},
%%            postcode={}, 
%%            state={},
%%            country={}}
%% \fntext[label3]{}

\title{Cross-Frequency Collaborative Training Network and Dataset for Semi-supervised  First Molar Root Canal Segmentation} %% Article title

%% use optional labels to link authors explicitly to addresses:
%% \author[label1,label2]{}
%% \affiliation[label1]{organization={},
%%             addressline={},
%%             city={},
%%             postcode={},
%%             state={},
%%             country={}}
%%
%% \affiliation[label2]{organization={},
%%             addressline={},
%%             city={},
%%             postcode={},
%%             state={},
%%             country={}}
\cortext[cor1]{Corresponding Author, Email: litao@nankai.edu.cn}
\cortext[cor2]{Co-first Authors, contribute equally}
\author[label1,label2]{Zhenhuan Zhou\corref{cor2}} %% Author name
\author[label3]{Yuchen Zhang\corref{cor2}}
\author[label1,label2]{Along He}
\author[label1]{Peng Wang}
\author[label4]{Xueshuo Xie}
\author[label1,label4]{Tao Li\corref{cor1}}

%% Author affiliation
\affiliation[label1]{organization={College of Computer Science, Nankai University},
            city={Tianjin},
            postcode={300350}, 
            country={China}}
\affiliation[label2]{organization={Key Laboratory of Data and Intelligent System Security, Ministry of Education},
            country={China}}
\affiliation[label3]{organization={Department of stomatology, Tianjin Union Medical Center},
            city={Tianjin},
            country={China}}
\affiliation[label4]{organization={Haihe Lab of ITAI},
            city={Tianjin},
            postcode={300459}, 
            country={China}}

%% Abstract
\begin{abstract}
%% Text of abstract
Root canal (RC) treatment is a highly delicate and technically complex procedure in clinical practice, heavily influenced by the clinicians' experience and subjective judgment. Deep learning has made significant advancements in the field of computer-aided diagnosis (CAD) because it can provide more objective and accurate diagnostic results. However, its application in RC treatment is still relatively rare, mainly due to the lack of public datasets in this field. To address this issue, in this paper, we established a First Molar Root Canal segmentation dataset called FMRC-2025. Additionally, to alleviate the workload of manual annotation for dentists and fully leverage the unlabeled data, we designed a Cross-Frequency Collaborative training semi-supervised learning (SSL) Network called CFC-Net. It consists of two components: (1) Cross-Frequency Collaborative Mean Teacher (CFC-MT), which introduces two specialized students (SS) and one comprehensive teacher (CT) for collaborative multi-frequency training. The CT and SS are trained on different frequency components while fully integrating multi-frequency knowledge through cross and full frequency consistency supervisions. (2) Uncertainty-guided Cross-Frequency Mix (UCF-Mix) mechanism enables the network to generate high-confidence pseudo-labels while learning to integrate multi-frequency information and maintaining the structural integrity of the targets. Extensive experiments on FMRC-2025 and three public dental datasets demonstrate that CFC-MT is effective for RC segmentation and can also exhibit strong generalizability on other dental segmentation tasks, outperforming state-of-the-art SSL medical image segmentation methods. Codes and dataset will be released.
\end{abstract}


\end{frontmatter}

%% Add \usepackage{lineno} before \begin{document} and uncomment 
%% following line to enable line numbers
%% \linenumbers

%% main text
%%

%% Use \section commands to start a section
\section{Introduction}
\label{sec1}
%% Labels are used to cross-reference an item using \ref command.
Apical periodontitis (AP) is an inflammatory condition affecting periapical tissues, with a global prevalence exceeding 50\% \cite{tiburcio2021global}. Root canal (RC) treatment is the primary treatment for AP, offering a means to preserve the essential function of affected teeth within the oral cavity \cite{leon2022prevalence}. However, the RC system is inherently complex and exhibits considerable variability in canal morphology across individuals, influenced by factors such as age and geographical region. This complexity makes the success of RC treatment heavily reliant on the clinician's expertise and subjective judgment, where even minor oversights can result in unpredictable outcomes or treatment failure \cite{meirinhos2020prevalence}. Therefore, there is an urgent need for an intelligent analytical approach to evaluate patients' RC systems prior to RC treatment. Such an approach would provide crucial insights into the number and morphology of canals, enabling clinicians to develop precise surgical plans while minimizing labor and financial costs. Achieving this objective necessitates accurate and reliable RC segmentation.

\begin{figure}[t]
	\centerline{\includegraphics[width=\columnwidth]{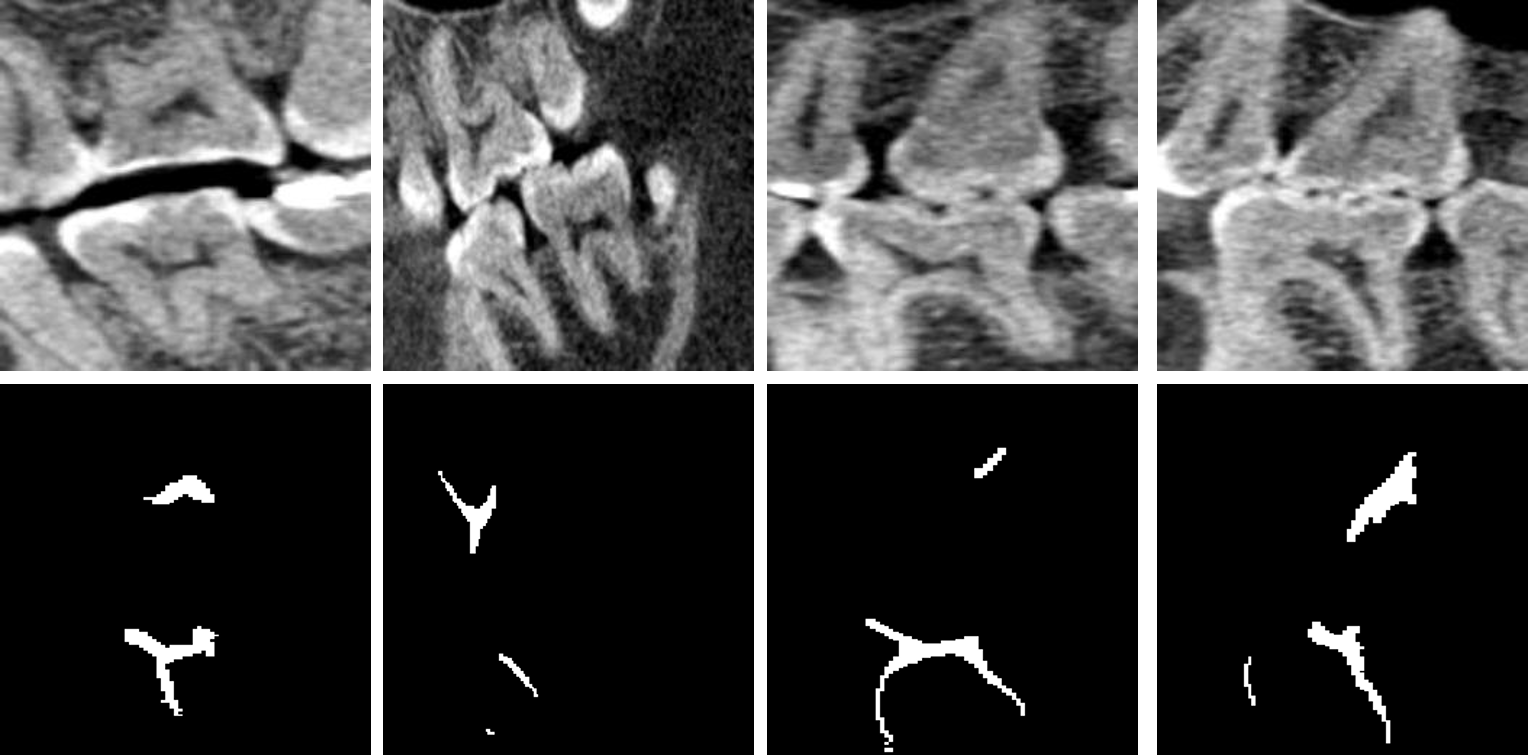}}
	\caption{Example images and their corresponding labels from FMRC-2025, they illustrate the intricate and highly variable morphology of root canals, which are small and difficult to distinguish. These complexities not only make the annotation process labor-intensive but also underscore the challenges in achieving accurate segmentation.}
	\label{fig1}
\end{figure}

Deep neural networks (DNN) have been widely employed in various medical image segmentation tasks, achieving remarkable success \cite{azad2024beyond}, \cite{liu2024rolling}, \cite{wu2023cross}, \cite{li2023can}, \cite{zhang2025large}, \cite{chen2024ma}. Inspired by these advancements, we aim to leverage DNN-based methods to address the challenges of RC segmentation. In recent years, several studies have explored the application of DNN in dental auxiliary diagnosis \cite{cui2019toothnet}, \cite{shi2022semantic}, \cite{cui2022fully}, \cite{chen2024cross}, \cite{jang2024fully}, and a number of dental datasets have been introduced \cite{cui2022ctooth}, \cite{zou2024teeth}, \cite{9557804}. However, the majority of existing research has focused on the segmentation and analysis of complete teeth within the oral cavity, paying little attention to the internal structures of teeth. Furthermore, there is a notable lack of publicly available datasets specifically designed for RC segmentation, along with the absence of effective models tailored to this task. These limitations have hindered the broader application of DNN in supporting periodontal diagnosis and treatment.

To address these issues, we introduce a First Molar Root Canal segmentation Cone Beam Computed Tomography (CBCT) dataset, FMRC-2025. This dataset contains 570 volumes from 235 clinical patients, among which the data of 30 patients were fully annotated by our expert annotation team, while the remaining cases were annotated using a human-machine hybrid method. Each volume includes pixel-level annotations of the RC for the upper and lower first molars (FM). In Section III, we provide a detailed description of FMRC-2025, including the rationale for selecting FM as the focus of our study, statistical details about the dataset, and the comprehensive processes involved in its collection, annotation, and establishment.

In clinical practice, unlabeled data are often more abundant and accessible than labeled data. Semi-Supervised Learning (SSL), known for their remarkable ability to utilize unlabeled data effectively, have become a preferred approach among researchers and clinicians \cite{he2023bilateral}. Two of the most representative SSL paradigms are Consistency Regularization \cite{tarvainen2017mean} and self-training \cite{lee2013pseudo}. Most existing SSL methods for medical image segmentation are built upon these foundational approaches or their combinations \cite{hang2020local}, \cite{wu2021semi}, \cite{zeng2023ss}, \cite{zhong2024semi}. The primary contributions of these methods lie in two key areas: the development of effective consistency constraints and the generation of high-quality, low-uncertainty pseudo-labels.

Although these methods have demonstrated good performance, they still exhibit some limitations. \textbf{First:} teacher networks often lack the ability for self-learning. As a result, when student networks make errors, teacher networks are unable to self-correct, leading to the accumulation of errors over time. \textbf{Second:} frequency domain information is critical for medical image segmentation \cite{zhou2023xnet}, \cite{wang2024fremim}, particularly for targets such as RC, which are small in size and characterized by abundant high-frequency details, as illustrated in Fig \ref{fig1}. However, most existing SSL methods for medical image segmentation primarily focus on feature learning in the spatial domain, neglecting the valuable information embedded in the frequency domain. \textbf{Third:} some prior methods incorporate the mix-up mechanism \cite{bai2023bidirectional}, \cite{shen2023co}, wherein patches with low uncertainty replace those with high uncertainty to generate new training samples and produce more reliable pseudo-labels. While this approach has proven effective, it suffers from a critical drawback, it compromises the structural integrity of lesions in the newly generated training images, potentially undermining the models' performance.

To address the aforementioned issues, we propose a SSL medical image segmentation network named CFC-Net. This network consists of two main components: the Cross-Frequency Collaborative Mean Teacher (CFC-MT) and the Uncertainty-guided Cross-Frequency Mix (UCF-Mix). In CFC-MT, we introduce a collaborative training framework that includes two specialized student (SS) networks and a comprehensive teacher (CT) network. The SS networks are designed to process low-frequency (LF) and high-frequency (HF) components of the entire-frequency (EF) image, respectively, while the CT network operates on the EF image. In the supervised path, the SS and CT networks are trained on labeled data using the segmentation loss. In the unsupervised path, the loss function consists of two components: the Full-frequency Consistency Supervision loss, which allows the CT network to guide the two SS networks from the perspective of the EF image, mitigating error accumulation during training; and the Cross-frequency Consistency Supervision loss between the two SS, which facilitates knowledge exchange between the two SS networks across different frequency domains, preventing them from working in isolation. The UCF-Mix mechanism employs a two-step cross-frequency mixing strategy. An integrated uncertainty map is generated from the outputs of the SS and CT networks, the top-k low-uncertainty foreground patches are then selected and mixed across the LF, EF, and HF images to create new training samples. These newly generated samples are subsequently fed back into the CFC-MT framework for further training. In Section IV, we will provide a detailed description of the structure of CFC-Net.

The key contributions of this paper can be summarized as follows:

\begin{itemize}
	\item[$\bullet$] We collected and annotated a CBCT dataset for FMRC segmentation, named FMRC-2025. To the best of our knowledge, this is the first and largest dataset in this field.
	\item[$\bullet$] We propose a collaborative training architecture called CFC-Net, which comprises two key components: CFC-MT and UCF-Mix. CFC-Net effectively leverages frequency domain information while maintaining sub-network divergence, ensuring high-quality pseudo-labels and preserving the structural integrity of the target.
	\item[$\bullet$] We conducted extensive experiments on FMRC-2025 and three additional public dental datasets. The results demonstrated that CFC-Net achieves excellent performance and robustness, exhibiting strong competitiveness against previous state-of-the-art (SOTA) SSL methods. Furthermore, ablation studies validated the effectiveness of each component within CFC-Net.
\end{itemize}

\section{RELATED WORKS}

\subsection{Deep Learning in Dental Medicine}
With the advancement of Deep Learning (DL), its applications in dentistry have become increasingly widespread, giving rise to a series of methods and datasets. In 2019, Cui et al. \cite{cui2019toothnet} proposed an automatic teeth identification and segmentation model for CBCT images. They used a framework supplemented with edge map features, achieving notable results on a small-scale private dataset. The Ctooth \cite{cui2022ctooth}, \cite{cui2022ctooth+} represent the first public CBCT tooth segmentation datasets, comprising 22 labeled cases and 111 unlabeled cases. Reference \cite{cui2022fully} introduced an intelligent system leveraging tooth centroids and skeletal information as guidance for automated instance segmentation of teeth and alveolar bone in CBCT images. Zou et al. \cite{zou2024teeth} introduced IO150K, the first intraoral image dataset, containing over 150,000 intraoral photographs. They also proposed a framework, TeethSEG, which integrates multi-scale aggregation and human prior knowledge to achieve tooth instance segmentation with good performance. DL has also been applied to RC analysis. For instance, Li et al. \cite{li2021agmb} developed a multi-stage network guided by the anatomical structure of tooth apices to evaluate RC treatment outcomes from X-ray images. Wang et al. \cite{wang2023root} employed a multi-task feature learning network, where the first stage performed tooth instance segmentation and the second stage segmented individual tooth RC. Notably, most prior studies have focused on segmenting entire teeth, with relatively limited research addressing the internal structure of RC. Even more regrettably, previous works utilizing DL for RC analysis did not provide publicly available RC segmentation datasets, which hinders further exploration of DL's potential applications in RC analysis. Therefore, public datasets remain crucial for advancing the application of DL in RC analysis.

\subsection{Semi-Supervised Medical Image Segmentation}
In recent years, numerous semi-supervised medical image segmentation methods have been proposed to fully leverage large volumes of unlabeled data. For instance, Wu et al. \cite{wu2023compete} introduced a competitive winning approach to generate high-quality pseudo-labels by comparing multiple confidence maps produced by different networks. References \cite{zhong2024semi} utilized a multi-branch architecture with a shared encoder and multiple decoders employing various attention mechanisms to enforce mutual consistency supervision. Liu et al. \cite{liu2022semi} proposed a combination of a shape-aware network and a shape-agnostic network to generate pseudo-labels and make effective use of unlabeled data. Zhou et al. \cite{zhou2023xnet} developed an X-shaped network for both fully supervised and semi-supervised segmentation tasks. This architecture integrates high and low-frequency features at the bottleneck layer and applies consistency constraints to the outputs of the respective frequency branches. It can be observed that most previous methods focus on semi-supervised feature learning in the spatial domain, neglecting the rich information in the frequency domain. XNet \cite{zhou2023xnet} introduced frequency domain information into semi-supervised learning, but it was designed based on a specific network architecture, making it non-transferable across different backbones.
\begin{figure*}[h]
	\centerline{\includegraphics[width=\linewidth]{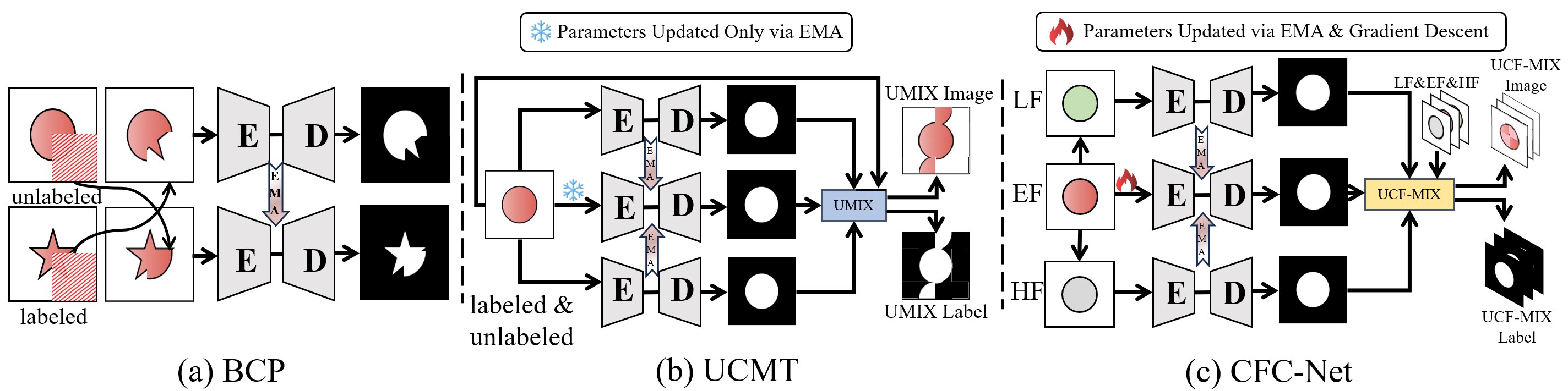}}
	\caption{A conceptual comparison of our method with two other mix-up mechanisms. EMA represents the exponential moving average. Our CFC-Net performs collaborative training across different frequency components, effectively utilizing each frequency component while maintaining high confidence and structural integrity of the labels.}
	\label{fig2}
\end{figure*}
\begin{figure*}[h]
	\centerline{\includegraphics[width=\linewidth]{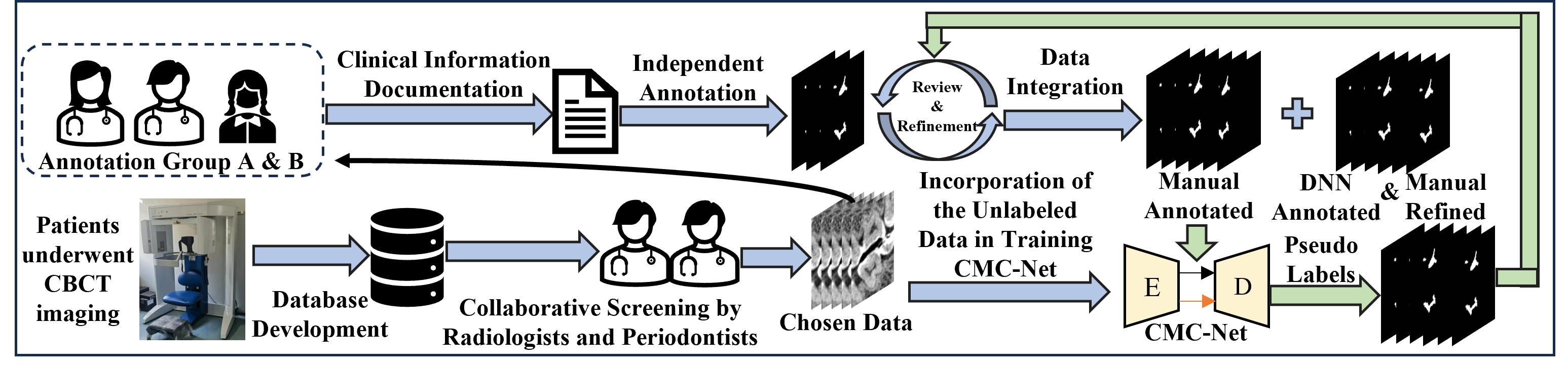}}
	\caption{The complete process diagram for the establishment of the FMRC-2025 dataset. The blue arrows represent manual operations, while the green arrows represent machine-assisted operations.}
	\label{fig3}
\end{figure*}

\subsection{Mix-Up Mechanism}
CutMix \cite{yun2019cutmix} is one of the most influential works in the mix-up mechanism. By combining patches from two different images, it enables the network to produce smoother decision boundaries and improve generalization. This approach has inspired many SSL medical image segmentation methods to build upon and extend its concepts. For example, Chen et al. \cite{chen2023magicnet} proposed a magic-cube partition and recovery method that preserves spatial positions while exchanging patches between paired 3D labeled and unlabeled data. This strategy allows the network to learn feature representations from these mixed cubes, thereby enhancing the quality of pseudo labels. In 2023, BCP \cite{bai2023bidirectional} was introduced (Fig. \ref{fig2} (a)), utilizing a bidirectional copy-paste mechanism within a MT framework to handle both labeled and unlabeled data, encouraging the unlabeled data to learn more comprehensive shared semantics from the labeled data. Shen et al. \cite{shen2023co} proposed a two-stage co-training architecture called UCMT. As shown in Fig. \ref{fig2} (b), it employs a Collaborative Mean Teacher (CMT) to encourage model divergence and perform collaborative training across sub-networks and uses Uncertainty-guided region mix to modify the input images, facilitating the production of high-confidence pseudo-labels by CMT. As shown in Fig. \ref{fig2} (c), inspired by UCMT, our CFC-Net also adopts a collaborative training approach but introduces several key distinctions. In CFC-MT, the three sub-networks are designed to learn from different frequency components, ensuring both network divergence and effective cross-frequency knowledge integration. Furthermore, in UCF-Mix, integrated uncertainty maps guide the bilateral mix of low-uncertainty patches among LF, EF, and HF components. This mechanism not only ensures low uncertainty in pseudo-labels but also enables the sub-networks to learn robust parameters for different frequency adjustments, all while preserving the structural integrity of the target.

\section{FMRC-2025 DATASET}
This study was approved by the Medical Ethics Committee of Tianjin Union Medical Center, confirming that all research content complies with the Declaration of Helsinki and the relevant regulations of the People's Republic of China on biological human experiments. Approval number: GZR2024027, approval date: February 26, 2024.

\subsection{Motivation}
Inspired by previous works\cite{li2021agmb}, \cite{wang2023root}, RC segmentation is often based on extracting the tooth of interest, so we selected the most representative teeth in the oral cavity, i.e., the FM, as the focus of the FMRC-2025 dataset. Once a DNN can accurately segment the RC of the FM, it can be conveniently extended to the segmentation of RC in other teeth. \textit{\textbf{Reason for choosing the FM:}} We chose the FM as the research focus, entirely based on the circumstances and needs of clinical dentisits. The FM are among the earliest permanent teeth to erupt, playing a critical role in mastication and maintaining occlusal stability. However, due to their complex anatomical structure, including numerous pits and fissures on the occlusal surface and poor self-cleaning capacity, they are highly susceptible to caries and pulp diseases. Consequently, in clinical practice, FM often require RC treatment \cite{chaleefong2021comparing}, \cite{ka2023cone}. Moreover, the RC of the FM are relatively dispersed and concealed, often resulting in inadequate filling or missed canals during clinical RC treatments, which heavily impacts the success of these procedures \cite{wu2021geometric}, \cite{wolcott20055}. In summary, accurate segmentation of the FMRC can provide valuable prior knowledge, enabling clinicians to formulate optimal treatment plans.
\begin{figure*}[t]
	\centerline{\includegraphics[width=\linewidth]{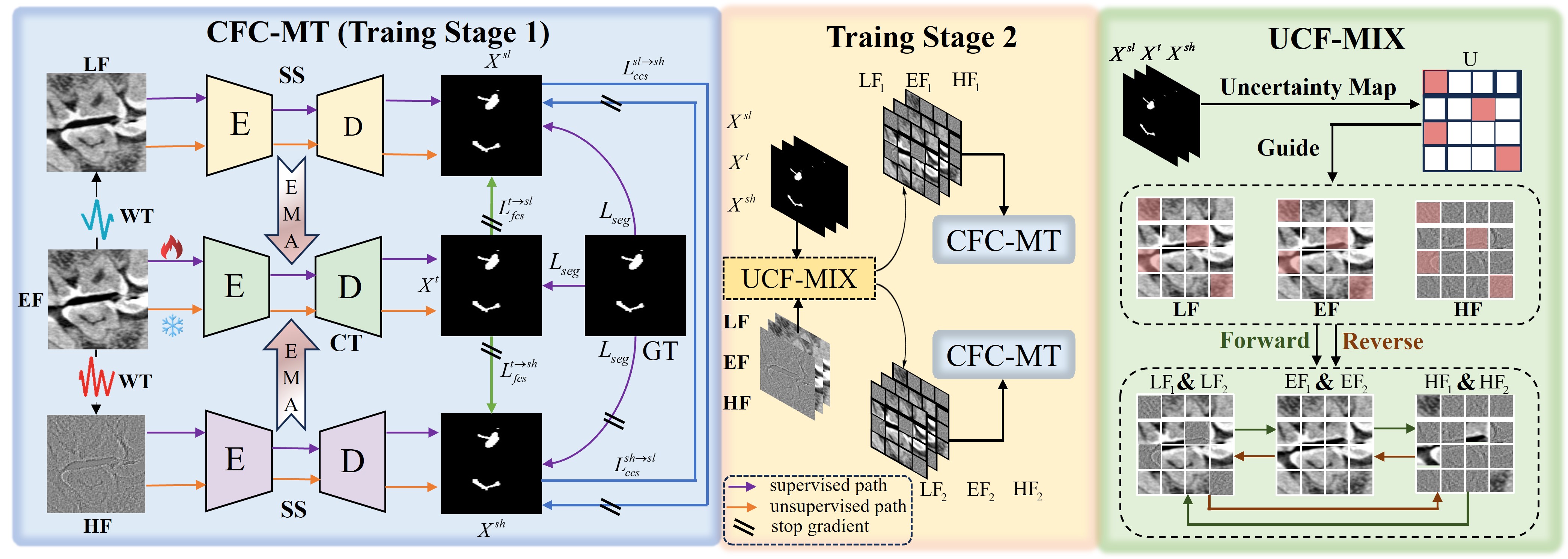}}
	\caption{Overall Structure of CFC-Net, EMA represents the Exponential Moving Average, and GT denotes Ground Truth. Given a batch of images, it first undergoes the first training stage through CFC-MT (blue), followed by the generation of new training samples using the UCF-Mix mechanism (yellow), and then the new samples are used to retrain CFC-MT (green).}
	\label{fig4}
\end{figure*}
\subsection{Collection and Annotation}
We collected CBCT images from 235 patients at the Tianjin Union Medical Center, encompassing a diverse range of ages, genders, and regions. The dataset includes 92 male and 143 female participants. Among them, 46 are adolescents aged 18 or younger (19.57\%), 130 are adults aged 19 to 40 (55.32\%), and 59 are middle-aged or elderly individuals over 40 (25.11\%), with the oldest participant being 68 years old. The overall process for constructing the FMRC-2025 dataset is illustrated in Fig. \ref{fig3}. Radiology and periodontology specialists initially selected suitable CBCT images from the database based on specific criteria: clear visibility of bilateral upper and lower FM, absence of significant lesions, and no history of RC treatment. For all selected images, patient de-identification was performed to preserve privacy, retaining only essential clinical information. The data were then randomly divided into two equal portions and assigned to two annotation groups for labeling. Specifically, we invited four clinically experienced periodontists and two computer science researchers to form two groups, referred to as Group A and Group B. The periodontists manually annotated the RC regions of the FM at the pixel level, while the computer science researchers performed necessary preprocessing and integrated the annotations. Inspired by \cite{zou2024teeth}, we adopted a human-machine hybrid annotation approach to reduce workload. Based on this, we employed CFC-Net as an assistant tool. Initially, each group manually annotated 15 data samples independently. Using these manually annotated data and the unlabeled data, we trained CFC-Net to generate pseudo-labels for the remaining images. The two groups then refined and adjusted the labels produced by the network, completing the human-machine hybrid annotation process. Finally, we separated the annotated regions of the four upper and lower FMs on both sides for each patient. Each patient corresponds to two volumes (left and right), resulting in a total of 570 volumes. The construction of the entire dataset spanned nearly 12 months. 

\section{METHODOLOGY}
\subsection{Overview}
In SSL for medical image segmentation, the training dataset typically consists of $M$ labeled samples and $N$ unlabeled samples, where $N \gg M$. The labeled images and their corresponding labels are defined as $ P_l \in \{x_i, y_i\}|_{i=1}^{M}$, and the unlabeled images are defined as $ P_u \in \{x_j\}|_{j=1}^{N}$. The entire training dataset for SSL is then represented as $ P_l \cup P_u$. Here, $x$ denotes the images, typically with a resolution of $\mathbb{R}^{H\times W \times C}$, and $y$ represents the corresponding labels with the same resolution, where pixel values range within $\{0, 1, \cdot\cdot\cdot, n \}$. $H$, $W$ and $C$ denote the height, width, and number of channels of the image, respectively, while $n$ is the number of segmentation classes.

As shown in Fig. \ref{fig4}, CFC-Net consists of two main components: the CFC-MT and the UCF-Mix mechanism. The training process of CFC-Net comprises three steps. Specifically, for a given input $\text{EF} \in \mathbb{R}^{H\times W \times C}$,  we first obtain its high-frequency component $\text{HF} \in \mathbb{R}^{H\times W \times C}$ and low-frequency component $\text{LF} \in \mathbb{R}^{H\times W \times C}$ via Wavelet Transform (WT) \cite{unser1995texture}. Then, LF and HF serve as inputs to the two SS networks within CFC-MT, while EF serves as the input to the CT network. The outputs from the two SS networks and the CT network undergo associated uncertainty estimation, generating an uncertainty map $U \in \mathbb{R}^{H\times W \times 1}$. Subsequently, guided by $U$, the UCF-Mix mechanism generates new mixed images. Finally, the training of CFC-MT is repeated using the newly generated mixed images. In the testing phase, only the CT network is required. Next, we will provide a detailed description of each component.

\subsection{Cross Frequency Collaborative Mean-Teacher}
\subsubsection{Motivation} Divergence between sub-networks is crucial for effective co-training \cite{shen2023co}, \cite{yu2019does}. Simultaneously, numerous studies have shown that fully utilizing frequency domain information can yield excellent results in medical image segmentation \cite{zhou2023xnet}, \cite{finder2025wavelet}, particularly for tasks like RC segmentation, where target shapes are complex and boundaries are easily confused. Building on this analysis, our CFC-MT adopts a structure consisting of two SS networks and one CT network. The HF and LF components are fed into the two SS networks, respectively, ensuring that they learn multi-frequency domain features while also maintaining sufficient divergence to prevent training degradation.The weights of the CT network are updated not only through the EMA of the SS networks, but also via direct training on EF images. This enhances the CT network’s self-training capability, allowing it to utilize labeled data more effectively and enabling self-correction. For unlabeled data, we introduce Cross-frequency
Consistency Supervision loss ($\mathcal L_{ccs}$) between the SS networks and Full-frequency Consistency Supervision loss ($\mathcal L_{fcs}$) between the CT and SS networks during training.

\subsubsection{Structural Details} As shown in the blue part of Fig \ref{fig4}, given an input image $\text{EF} \in \mathbb{R}^{H\times W \times C}$, the first step is to apply a WT, as described in Equation (\ref{eq1}), to extract its frequency components: $LL, HL, LH$ and $HH$, where $\psi_{F}$ denotes the wavelet coefficients. In our experiments we use db2, $(n, m)$ represent the pixels in the EF image, while $(x, y)$ represent the coordinates in the domain of $\psi_{F}$.

\begin{equation}
LL, (LH, HL, HH) = \frac{1}{\sqrt{2}} \sum_{n,m} \text{EF}(n,m) \psi_{F}(x-n, y-m)
\label{eq1}
\end{equation}
The LF corresponds to the $LL$ component, while the HF is the sum of $HL$, $LH$ and $HH$. The CT and SS network structures are identical. For labeled data, all networks are trained under supervision using the labels. For unlabeled data, we use two loss functions: the $\mathcal L_{fcs}$ and the $\mathcal L_{ccs}$. The loss function $\mathcal L$ for the entire training process consists of two parts, i.e., $\mathcal L_{sup}$ and $\mathcal L_{unsup}$, as shown in Equation \ref{eq3}.

\begin{equation}
\mathcal L = \mathcal L_{sup} + \lambda \mathcal L_{unsup}
\label{eq3}
\end{equation}
where $\lambda$ is a regularization term used to control the weight of the unsupervised loss, defined by a Gaussian warm-up function, as shown in Equation (\ref{eq4})
\begin{equation}
\lambda(t) = \lambda_{\text{max}} \cdot e^{-5 \left(1 - \frac{t}{t_{\text{m}}}\right)^2}
\label{eq4}
\end{equation}
where $t$ denotes the current iteration and $\lambda(t)$ represents the value of $\lambda$ at iteration $t$, $t_m$ is the total number of iterations, and $\lambda_{max}$ is a hyperparameter representing the maximum value of $\lambda$.

\textbf{Supervised Path:} The SS networks and the CT network are trained on labeled data. For paired LF, EF, and HF components, all are trained using the same label. The loss function $\mathcal{L}_{sup1}$ for the first supervised training stage is defined in Equation (\ref{eq5}).
\begin{equation}
\mathcal L_{sup1} = \sum_{i=1}^{m}(\mathcal L_{seg}(x_i^{sl}, y_i) + \mathcal L_{seg}(x_i^{sh}, y_i) + \mathcal L_{seg}(x_i^{t}, y_i))
\label{eq5}
\end{equation}
where $\mathcal{L}_{\text{seg}}$ represents the segmentation loss. The terms $x_i^{sl}$, $x_i^{sh}$, and $x_i^{t}$ denote the outputs of the low-frequency SS network, the high-frequency SS network, and the CT network, respectively, while $y$ represents the corresponding labels.

\textbf{Unsupervised Path:} In the unsupervised path, the parameters of the CT network are updated using the combined EMA of the two SS networks, as defined in Equation (\ref{eq6}).

\begin{equation}
\theta_t = \alpha\theta_{t-1} + \beta (1-\alpha)\theta_{sl}^{t} + (1-\beta)(1-\alpha)\theta_{sh}^{t}
\label{eq6}
\end{equation}
where $\alpha$ and $\beta$ are smoothing coefficients that control the EMA update rate and the relative contributions of the high-frequency and low-frequency SS networks to the EMA, respectively. $\theta_*^{t}$ denotes the weights of the corresponding network at the $t$-th iteration. In the first unsupervised training stage, the loss function $\mathcal{L}_{unsup1}$ comprises two components, as defined in Equation (\ref{eq7}).

\begin{equation}
\mathcal L_{unsup1} = \mathcal L_{fcs} + \mathcal L_{ccs}
\label{eq7}
\end{equation}
The $\mathcal L_{fcs}$ supervises the outputs of both SS networks simultaneously using pseudo-labels generated by the CT network. The primary goal of $\mathcal L_{fcs}$ is to enable the CT network to guide and constrain the training processes of the SS networks from a full-frequency perspective, thereby preventing errors and the accumulation of mistakes during training. As shown in Equations (\ref{eq8}) and (\ref{eq9}), the formulation of $\mathcal L_{fcs}$ is provided using $\mathcal{L}_{fcs}^{t \rightarrow sl}$ as an example, where $y_t$ denotes the pseudo-labels generated by the CT network.
\begin{equation}
    \mathcal L_{fcs}^{t \rightarrow sl} = \sum_{j=1}^{n}( \mathcal L_{seg}(x_{sl}^{j}, y_{t}^{j}))
\label{eq8}
\end{equation}

\begin{equation}
    \mathcal L_{fcs} = \frac{1}{2}(\mathcal L_{fcs}^{t \rightarrow sl} + \mathcal L_{fcs}^{t \rightarrow sh})
\label{eq9}
\end{equation}
In $\mathcal{L}_{ccs}$, the two SS networks generate pseudo-labels for each other to enable mutual supervision. This consistency mechanism allows the SS networks to learn collaboratively, leveraging their respective strengths to compensate for weaknesses, correct errors, and avoid isolated learning. As shown in Equations (\ref{eq10}) and (\ref{eq11}), the formulation of $\mathcal{L}_{ccs}$ is provided using $\mathcal{L}_{ccs}^{sl \rightarrow sh}$ as an example, where $y_{sl}$ denotes the pseudo-labels generated by the low-frequency SS network.

\begin{equation}
    \mathcal L_{ccs}^{sl \rightarrow sh} = \sum_{j=1}^{n}( \mathcal L_{seg}(x_{sh}^{j}, y_{sl}^{j}))
\label{eq10}
\end{equation}

\begin{equation}
    \mathcal L_{ccs} = \frac{1}{2}(\mathcal L_{ccs}^{sl \rightarrow sh} + \mathcal L_{ccs}^{sh \rightarrow sl})
\label{eq11}
\end{equation}

\subsection{Uncertainty-guided Cross-Frequency Mix mechanism}
\subsubsection{motivation} The motivation behind the design of UCF-Mix is to develop a mix-up method tailored for cross-frequency collaborative training. This approach aims to maintain high-confidence pseudo-labels while preserving the structural integrity of segmentation targets. UCF-Mix facilitates the bilateral mix-up of high-confidence patches across LF, EF, and HF components. By explicitly incorporating cross-frequency information into the sub-networks, this method enhances the robustness of the CT to various frequency components and prevents the SS networks from being confined to their respective frequency domains.

\subsubsection{Structural Details}
We illustrate the UCF-Mix generation process using the supervised mixed samples as an example; the procedure for generating new unlabeled samples is similar. After obtaining the outputs $x_{sl}$, $x_{sh}$, and $x_t$ from the SS and CT networks, we first compute their joint probability distribution $P$, as defined in Equation (\ref{eq12}). Subsequently, the uncertainty map $U$ is calculated using Equation (\ref{eq13}), where $c$ represents the number of segmentation classes, and $\epsilon$ is a small parameter introduced to prevent logarithmic calculations from approaching zero.
\begin{equation}
P = \frac{1}{3}(\text{Softmax}(x_{sl})+ \text{Softmax}(x_{sh}) + \text{Softmax}(x_{t}))
\label{eq12}
\end{equation}
\begin{equation}
U = -\sum_{c}(P_{c}\cdot log(P_{c} + \epsilon))
\label{eq13}
\end{equation}
As shown in the green part of Fig .\ref{fig4}, the uncertainty map $U$ is divided into $k$ patches and ranked. The top 25\% high-confidence foreground patches are selected for the bilateral mix-up. In the first round, the selected patches are mixed in a forward sequence of $sl \rightarrow t \rightarrow sh \rightarrow sl$, resulting in the first-round mixed training samples: $\text{LF}_1$, $\text{EF}_1$, and $\text{HF}_1$. In the second round, the same patches are mixed in a reverse sequence of $sh \rightarrow t \rightarrow sl \rightarrow sh$, producing the second-round mixed training samples: $\text{LF}_2$, $\text{EF}_2$, and $\text{HF}_2$. These generated samples are subsequently fed into the CFC-MT for the second stage of training, as shown in the yellow part of Fig .\ref{fig4}. The supervised and unsupervised loss functions for this stage are denoted as $\mathcal{L}_{sup2}$ and $\mathcal{L}_{unsup2}$, respectively, which follow a similar form to $\mathcal{L}_{sup1}$ and $\mathcal{L}_{unsup1}$ (It should be noted that the newly generated samples from the two rounds are each used for one training iteration). The total loss for the entire training process, $\mathcal{L}_{sup}$ and $\mathcal{L}_{unsup}$, is the sum of the losses from both stages, as defined in Equation (\ref{eq14}).
\begin{equation}
    \begin{aligned}
    \mathcal L_{unsup} &= \mathcal L_{unsup1} + \mathcal L_{unsup2} \\
    \mathcal L_{sup} &= \mathcal L_{sup1} + \mathcal L_{sup2}
    \end{aligned}
\label{eq14}
\end{equation}
It is worth noting that UCF-Mix only mixes patches at the same locations across different frequency components. As a result, it does not require any modifications to the training labels, thereby preserving the structural integrity of the target.

\section{EXPERIMENTS}
\subsection{Dataset}
In addition to FMRC-2025, we conducted experiments on three publicly available dental datasets to validate the effectiveness of CFC-Net. \textbf{CTooth} \cite{cui2022ctooth}, \cite{cui2022ctooth+} is a publicly available 3D CBCT tooth segmentation dataset consisting of 22 labeled and 111 unlabeled images. The 22 labeled images were split into two subsets: 16 images, together with all unlabeled images, constituted the training set, while the remaining 6 images served as the test set. \textbf{TDD} \cite{9557804} dataset is a 2D X-ray panoramic radiography image dataset consists of 1000 images. We randomly split the dataset into training set and test set with an 8:2 ratio. \textbf{NKUT} \cite{zhou2024nkut} is a dataset specifically designed for segmenting pediatric mandibular wisdom teeth (MWT) from CBCT images. This dataset contains 133 samples with three categories: bilateral MWT germs, second molars (SM), and partial alveolar bone (AB). For FMRC-2025 and NKUT, We randomly selected 80\% of the data as the training set and 20\% as the test set. 

\subsection{Implementation Details}
During training, we applied random horizontal and vertical flipping, as well as random rotation, as data augmentation strategies. For 3D dataset, we extract their 2D slices for training. The resolution of all images and labels was resized to $256 \times 256$. We utilized a UNet \cite{ronneberger2015u} with a VGG16 \cite{simonyan2014very} backbone as the training network. The encoder channel numbers were set to $[64, 128, 256, 512]$. The optimizer was Adam, with an initial learning rate of 0.0001, and the learning rate decay strategy followed a "poly" schedule. The epochs was set to 300. For UCMT \cite{shen2023co}, due to its two-step training process within each iteration, the epochs are set to 150. Similarly, the total epochs for CFC-Net were set to 100. The batch size was set to 12, consisting of 6 labeled and 6 unlabeled samples, K is set to 16, $\alpha$ and $\beta$ is set to 0.99, $\lambda_{max}$ is set to 0.1. The loss function used in the experiments is a combination of Cross-Entropy Loss and Dice Loss \cite{milletari2016v}. All experiments were implemented using PyTorch and conducted on two NVIDIA GeForce 3090 GPUs. The evaluation metrics included Mean Absolute Error (MAE), Recall, Dice Similarity Coefficient (DSC), Intersection-over-Union (IoU), 95\% Hausdorff Distance (HD95), and Average Surface Distance (ASD).

\subsection{Experimental Results}
To demonstrate the effectiveness of our method in SSL medical image segmentation tasks, we compared it with several previous SOTA methods, including MT \cite{tarvainen2017mean}, DTC \cite{luo2021semi}, CPS \cite{chen2021semi}, URPC \cite{luo2021efficient}, SASSNet \cite{li2020shape}, SSNet \cite{wu2022exploring}, UAMT \cite{yu2019uncertainty}, UCMT \cite{shen2023co}, BCP \cite{bai2023bidirectional}, and ABD \cite{chi2024adaptive}.

\subsubsection{Results on the FMRC-2025}
\begin{figure}[t]
	\centerline{\includegraphics[width=\columnwidth]{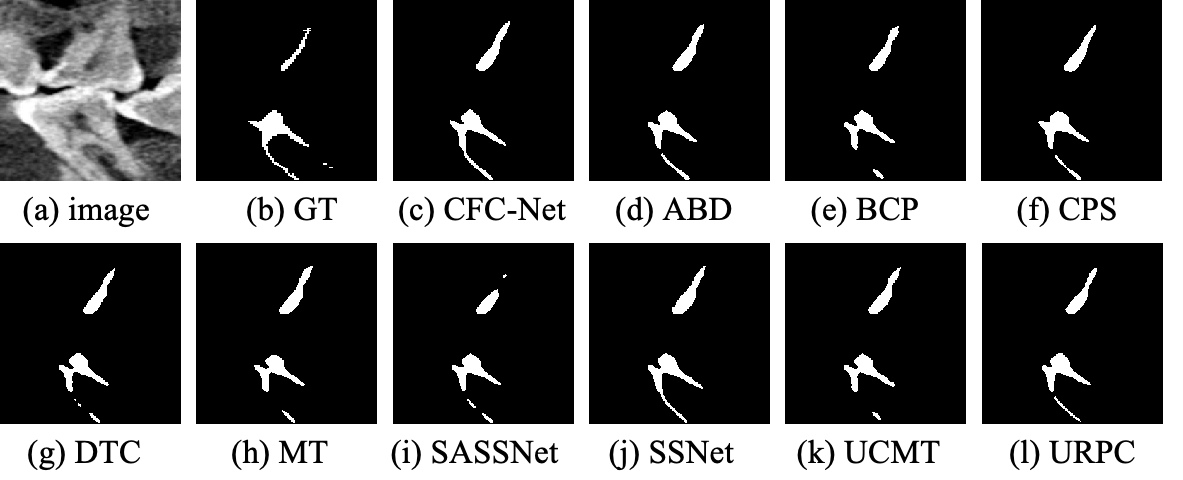}}
	\caption{Qualitative comparison of CFC-Net with other SOTA SSL networks on the FMRC-2025 dataset. "Image" represents the original input image, while "GT" denotes the ground truth labels.}
	\label{fig5}
\end{figure}

 As shown in Fig. \ref{fig5}, the qualitative comparisons of CFC-Net with other SOTA SSL medical image segmentation methods are presented. It can be observed that CFC-Net closely approximates the ground truth (b), achieving accurate and complete segmentation, even in the fine, intricate regions deep within the RC. In contrast, other networks exhibit omissions in identifying these subtle structures. This underscores the critical importance of UCF-Mix in preserving the integrity of segmentation labels. Table \ref{tab1} presents the quantitative results of the experiments, we conducted experiments using 10\% and 20\% of the labeled data, respectively. The results show that CFC-Net outperforms existing SOTA methods across most evaluation metrics. Notably, compared to recent SOTA SSL medical segmentation methods that utilize mix-up mechanisms, CFC-Net exhibits advantages in the RC segmentation task.
 \begin{table}[h]
\setlength{\tabcolsep}{2.5pt} % 设置单元格内左右间距为 4pt
        \centering
        \small
        \caption{The quantitative experimental results on the FMRC-2025 dataset. Each experiment was conducted five times, and the average results are reported. Bold text indicates the best performance, while underlined text denotes the second-best performance. $\text{P}_l$ denotes the ratio of labeled data.}
        \label{tab1}
        \begin{tabular}{c|c|cccc}
            \hline
             $\text{P}_l$& Model & DSC $\uparrow$ & IOU $\uparrow$ & HD95 $\downarrow$ & ASD $\downarrow$ \\
            \hline
            \hline
             \multirow{11}{*}{20\%}&MT $\text{NIPS'17} $ & 59.13 & 45.29 & 35.89 & 11.74 \\
            &UAMT $\text{(MICCAI'19)} $ & 60.04 & 46.05 & \underline{32.23} & 10.01 \\
            &SASSNet $\text{(MICCAI'20)} $& 59.02 & 45.12& 32.58 & 10.77 \\
             &CPS $\text{(CVPR'21)} $ & 59.33 & 45.38 & 33.31 & 11.33 \\
             &DTC $\text{(AAAI'21)} $ & 58.85 & 45.20 & 32.92 & 10.37 \\
            &URPC $\text{(MICCAI'21)} $ & 59.83 & 45.86 & 32.09 & 9.92\\
             &SSNet $\text{(MICCAI'22)} $ & 59.12 &  45.31& 32.61 & 10.13\\
             &UCMT $\text{(IJCAI'23)} $ & \underline{60.51} & \underline{46.52} & 32.36 & 10.34\\
             &BCP $\text{(CVPR'23)} $ & 59.82 & 45.87 & 33.59 & \underline{9.83}\\
             &ABD $\text{(CVPR'24)} $ & 59.26 & 45.62 & 33.34 & 10.92 \\
             \hline
             &CFC-Net (ours)& \textbf{60.71} & \textbf{46.69} & \textbf{31.91} & \textbf{9.71} \\
            \hline
            \hline
             \multirow{11}{*}{10\%}&MT & 55.21 & 41.82 & 40.24 & 11.95 \\
            &UAMT & 55.38 & 42.06 & 39.80 & 12.91 \\
            &SASSNet& 52.32 & 40.88& 40.71 & \textbf{11.33} \\
             &CPS & 55.99 & 42.26 & 41.46 & 13.24 \\
             &DTC & 53.87 & 40.93 & \underline{39.14} & 12.04 \\
            &URPC & 55.38 & 42.06 & 40.61 & 11.69\\
             &SSNet & 55.12 &  41.87 & 42.88 & 13.52\\
             &UCMT  & \underline{56.08} & \underline{42.39} & 39.80 & 12.91\\
             &BCP & 55.11 & 41.41 & 44.23 & \underline{15.33}\\
             &ABD & 55.64 & \underline{42.39} & 41.67 & 12.73 \\
             \hline
             &CFC-Net (ours)& \textbf{56.39} & \textbf{42.69} & \textbf{38.43} & \underline{11.53} \\
            \hline
        \end{tabular}
			
\end{table}

\subsubsection{Results on the TDD} 
\begin{figure}[t]

\centerline{\includegraphics[width=\columnwidth]{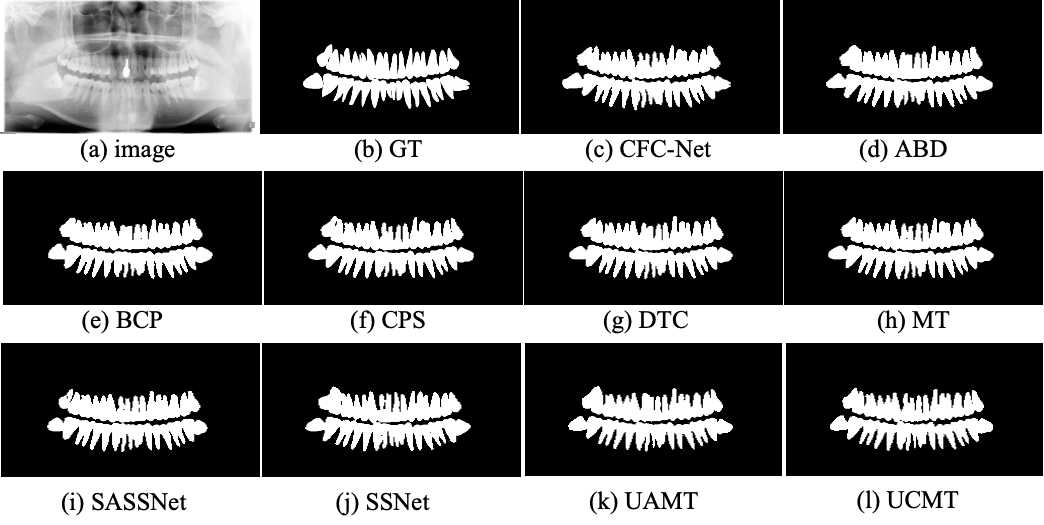}}
	\caption{Qualitative comparison of CFC-Net with other SOTA SSL networks on the TDD dataset.}
	\label{fig6}
\end{figure}

Fig. \ref{fig6} illustrates the visualization results of CFC-Net compared to other SOTA SSL networks on the TDD dataset. It is evident that, compared to other SOTA SSL networks, the segmentation results produced by CFC-Net exhibit the clearest boundaries. CFC-Net effectively learns detailed features of the tooth root region from low-contrast X-ray images. This highlights CFC-Net’s ability to effectively learn rich high-frequency edge information and low-frequency texture information through cross-frequency features. Table \ref{tab2} presents the quantitative results of the experiments, showing that CFC-Net achieves competitive performance. These findings clearly demonstrate the advantages of CFC-Net in semi-supervised 2D X-ray tooth segmentation tasks.
\begin{table}[h]
\setlength{\tabcolsep}{4.5pt} % 设置单元格内左右间距为 4pt
        \centering
        \small
        \caption{The quantitative experimental results on the TDD dataset. Each experiment was conducted five times, and the average results are reported. $\text{P}_l$ denotes the ratio of labeled data.}
        \label{tab2}
        \begin{tabular}{c|c|cccc}
            \hline
             $\text{P}_l$&Model & MAE $\downarrow$ & Recall $\uparrow$ & DSC $\uparrow$ & IOU $\uparrow$ \\
            \hline
            \hline
             \multirow{11}{*}{20\%}&MT & \underline{2.24} & \underline{90.83} & 89.93 & 82.14\\
            &UAMT  & 2.26 & 89.71 & 89.92 & 82.11\\
            &SASSNet & 2.35 & 89.54 & 89.42 & 81.45\\
             &CPS & 2.28 & 90.17 & 89.81 & 81.89 \\
             &DTC & 2.29 & 89.41 & 89.86 & 82.01\\
            &URPC & 2.29 & 89.86 & 89.29 & 81.60\\
             &SSNet & 2.32 &  89.54 & 89.15 & 81.33 \\
             &UCMT & \textbf{2.19}  & 90.19 & 89.18 & 82.02\\
             &BCP & 2.59 & 84.93 & 85.37 & 76.55\\
             &ABD & \underline{2.24} & \underline{90.83} & \underline{89.94} & \textbf{82.22} \\
             \hline
             &CFC-Net (ours)& \underline{2.24} & \textbf{91.29} & \textbf{89.96}& \underline{82.16}\\
            \hline
            \hline
            \multirow{11}{*}{10\%}&MT  & 2.43 & 88.71 & \underline{89.07} & 80.82\\
            &UAMT  & 2.40 & 89.15 & 88.76 & 80.61\\
            &SASSNet & 2.58 & 88.41 & 87.62 & 79.04\\
             &CPS & 2.41 & 88.83 & 88.91 & 80.74 \\
             &DTC & 2.42 & 88.54 & 88.98 & 80.71\\
            &URPC & 2.36 & 89.15 & 88.22 & 80.27\\
             &SSNet  & 2.44 &  90.54 & 86.49 & 78.45 \\
             &UCMT & \textbf{2.31}  & \textbf{89.96} & 88.97 & \underline{80.94}\\
             &BCP & 2.76 & 85.93 & 85.68 & 76.57\\
             &ABD & 2.39 & 89.68 & 89.04 & 80.88 \\
             \hline
             &CFC-Net (ours)& \underline{2.34} & \underline{89.84} & \textbf{89.11} & \textbf{80.97} \\
            \hline

        \end{tabular}
			
\end{table}

\subsubsection{Results on the CTooth}
\begin{figure}[t]
	\centerline{\includegraphics[width=\columnwidth]{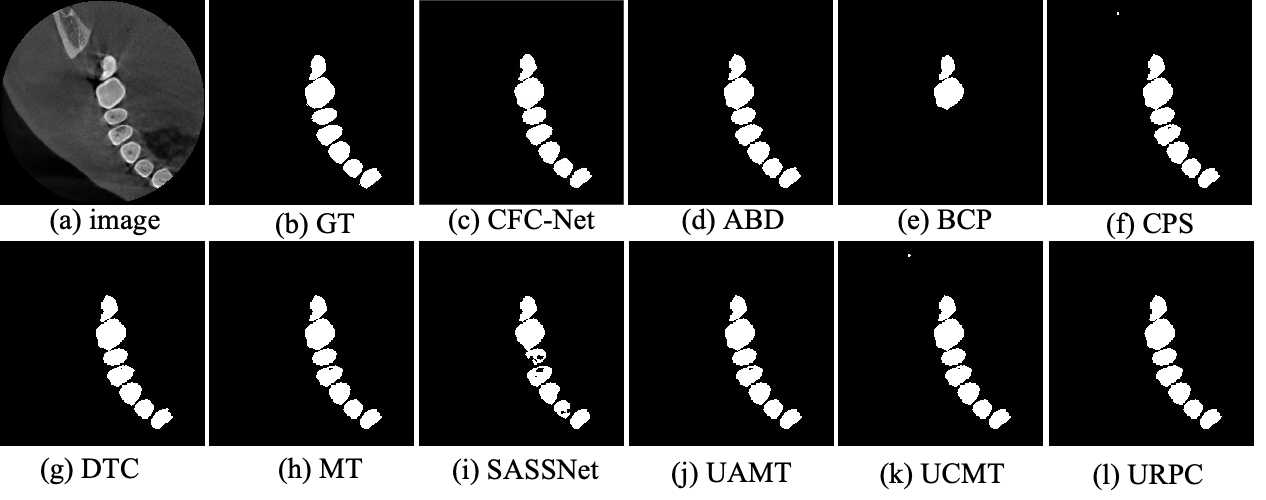}}
	\caption{Qualitative comparison of CFC-Net with other SOTA SSL networks on the CTooth dataset.}
	\label{fig7}
\end{figure}

Fig. \ref{fig7} illustrates the qualitative visualization results of the experiments on the CTooth dataset. It is evident that CFC-Net achieves the best performance, producing results that are the closest to the ground truth. Compared to other methods, CFC-Net provides a clearer recognition of the tooth contours, while maintaining the integrity of the tooth texture without any missing teeth. Table \ref{tab3} presents the quantitative results of the experiments, we compared the performance using all labeled data (16 images) and half of the labeled data (8 images). As can be seen from the results, CFC-Net achieves the best performance in DSC, IoU and ASD metrics. The results from these experiments provide substantial evidence that CFC-Net performs excellently in adult CBCT tooth segmentation, highlighting its versatility to various dental segmentation tasks.
\begin{table}[t]
\setlength{\tabcolsep}{5pt} % 设置单元格内左右间距为 4pt
        \centering
        \small
        \caption{The quantitative experimental results on the CTooth dataset. Each experiment was conducted five times, and the average results are reported. $\text{P}_l$ denotes the number of labeled data.}
        \label{tab3}
        \begin{tabular}{c|c|cccc}
            \hline
             $\text{P}_l$&Model & DSC $\uparrow$ & IOU $\uparrow$ & HD95 $\downarrow$ & ASD $\downarrow$ \\
            \hline
            \hline
             \multirow{11}{*}{16}&MT & 85.43 & 76.66 & 9.01 & 2.63\\
            &UAMT  & 85.57 & 76.91 & 8.94 & \underline{2.21}\\
            &SASSNet  & 84.29 & 75.62 & 11.20 & 3.49\\
             &CPS & 85.87 & 77.31 & 11.61 & 4.16 \\
             &DTC & 85.42 & 76.76 & 9.68 & 2.85\\
            &URPC & 85.91 & 76.87 & \textbf{8.69} & 2.36\\
            &SSNet  & 85.10 &  76.24 & 10.89 & 3.87 \\
             &UCMT & \underline{86.08} & \underline{77.68} & 9.61 & 3.51\\
             &BCP & 85.86 & 75.37 & 11.18 & 4.21 \\
             &ABD & 84.78 & 76.03 & 9.93 & 3.16 \\
             \hline
             &CFC-Net (ours) & \textbf{86.43} & \textbf{78.11} & \underline{8.88} & \textbf{2.15} \\
            \hline
            \hline
            \multirow{11}{*}{8}&MT & 82.78 & 73.05 & \underline{9.66} & 3.04\\
            &UAMT & 82.54 & 72.76 & 10.65 & 3.88\\
            &SASSNet & 80.44 & 70.66 & 12.61 & 4.21\\
             &CPS  & 82.31 & 72.68 & 15.06 & 6.28 \\
             &DTC  & 82.52 & 72.52 & 10.16 & \underline{3.03}\\
            &URPC  & \underline{83.04} & \underline{73.39} & 10.18 & 4.12 \\
            &SSNet & 81.30 &  71.39 & 13.18 & 5.17 \\
             &UCMT & 82.97 & 73.37 & 9.79 & 3.39\\
             &BCP & 81.88 & 71.95 & 13.87 & 4.65 \\
             &ABD & 81.05 & 71.01 & 12.78 & 4.62 \\
             \hline
             &CFC-Net (ours) & \textbf{83.57} & \textbf{74.28} & \textbf{9.54} & \textbf{2.94} \\
            \hline
        \end{tabular}
			
\end{table}

\subsubsection{Results on the NKUT}
\begin{figure}[t]
	\centerline{\includegraphics[width=\columnwidth]{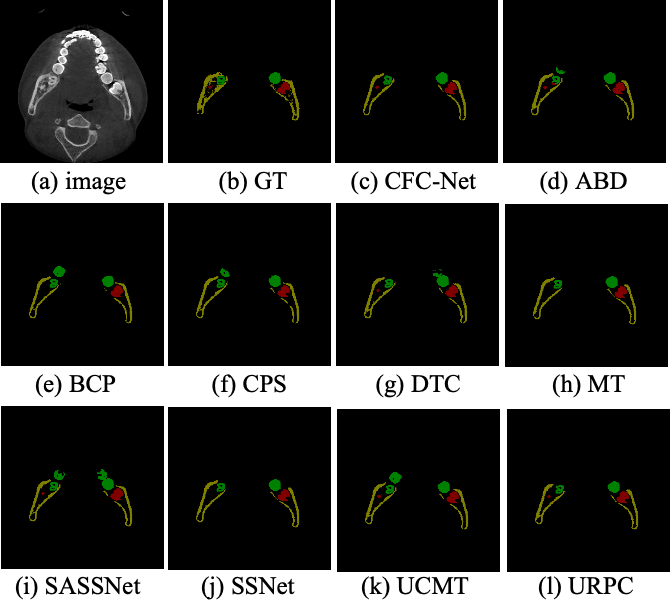}}
	\caption{Qualitative comparison of CFC-Net and other SOTA SSL networks on the NKUT dataset. Red, green and yellow represent MWT, SM and AB, respectively.}
	\label{fig9}
\end{figure}

The challenges associated with the NKUT dataset include multi-scale issues and feature confusion. Fig. \ref{fig9} presents the visual comparison results of the experiments. It can be observed that the segmentation results of CFC-Net exhibit clear boundaries between the teeth and bones, with no confusion between different teeth. In contrast, many other SOTA methods confuse the FM with the SM. Table \ref{tab5} presents the quantitative results of the experiments on the NKUT dataset, we also conducted comparative experiments using 20\% and 10\% of the labeled data. CFC-Net achieved the highest ranking and attained the optimal overall results. This demonstrates that CFC-Net benefits from its robust cross-frequency feature learning and reliable high-confidence pseudo-label generation capabilities, enabling it to effectively handle both large and small targets. Additionally, experiments on the NKUT dataset further highlight the strengths of CFC-Net in pediatric CBCT tooth segmentation tasks.

\begin{table}[h]
\setlength{\tabcolsep}{5pt} % 设置单元格内左右间距为 4pt
        \centering
        \small
        \caption{The quantitative experimental results on the NKUT dataset. Each experiment was conducted five times, and the average results are reported. $\text{P}_l$ denotes the ratio of labeled data.}
        \label{tab5}
        \begin{tabular}{c|c|cccc}
            \hline
             $\text{P}_l$&Model & DSC $\uparrow$ & IOU $\uparrow$ & HD95 $\downarrow$ & ASD $\downarrow$\\
            \hline
            \hline
             \multirow{11}{*}{20\%}&MT & 79.58 & 73.28 & \underline{22.67} & \textbf{5.83}\\
            &UAMT & 79.11 & 74.21 & 24.16 & 6.71 \\
            &SASSNet & 74.97 & 68.60 & 23.74 & 8.93 \\
             &CPS & 79.19 & 72.98 & 25.54 & 6.38 \\
             &DTC & 75.03 & 68.70 & 25.85 & 8.04 \\
            &URPC & 77.27 & 70.79 & 23.52 & 8.18 \\
             &SSNet & 75.39 & 69.01 & 26.66 & 7.33 \\
             &UCMT & \underline{80.45} & \underline{74.25} & 24.16 & 6.71 \\
             &BCP & 73.26 & 67.30 & 30.33 & 5.98 \\
             &ABD & 76.91 & 70.44 & 25.53 & 8.84 \\
             \hline
             &CFC-Net (ours) & \textbf{81.06} & \textbf{74.73} & \textbf{20.21} & \underline{5.91} \\
            \hline
            \hline
            \multirow{11}{*}{10\%}&MT & 78.42 & 72.11 & 28.34 & \underline{8.75}\\
            &UAMT & 78.65 & 72.28 & 27.18 & 7.99 \\
            &SASSNet & 71.42 & 65.18 & 32.37 & 9.36 \\
             &CPS & 78.44 & 71.75 & 27.13 & 8.99 \\
             &DTC & 73.01 & 66.71 & 33.54 & 8.79 \\
            &URPC & 76.47 & 70.14 & \underline{26.62} & 9.15 \\
             &SSNet & 69.42 & 63.23 & 36.57 & 9.88 \\
             &UCMT & \underline{79.02} & \underline{72.68} & \underline{26.62} & 9.14 \\
             &BCP & 61.66 & 57.41 & 41.45 & 9.54 \\
             &ABD & 75.40 & 69.07& 27.91 & 8.93 \\
             \hline
             &CFC-Net (ours) & \textbf{79.48} & \textbf{73.14} & \textbf{25.91} & \textbf{8.69} \\
             \hline
        \end{tabular}
			
\end{table}

\subsection{Ablation Studies}
We conducted extensive ablation studies on the FMRC-2025 dataset to verify the effectiveness of each component in CFC-Net. The quantitative results of all experiments are reported in Table \ref{tab6}, with each experiment being repeated three times to report the average results.
\begin{table}[t]
\setlength{\tabcolsep}{1.5pt} % 设置单元格内左右间距为 4pt
        \centering
        \small
        \caption{The quantitative results of the ablation experiments on the FMRC-2025 dataset (20\% labeled), where BL, CT, $\text{L}_s$ and $\text{H}_s$ represent the Baseline, comprehensive teacher, low and high frequency SS networks, respectively. $\checkmark$ indicates inclusion and $\circ$ indicates exclusion.}
        \label{tab6}
        \begin{tabular}{|c||cccccc||ccc|}
            \hline
              & CT & $\text{L}_s$ & $\text{H}_s$ & $\mathcal L_{fcs}$ & $\mathcal L_{ccs}$ & UCF-Mix & DSC$\uparrow$ & IOU$\uparrow$ & HD95$\downarrow$ \\
            \hline
            BL & $\circ$ & $\circ$ & $\circ$ & $\circ$ & $\circ$ & $\circ$ & 59.13 \vline & 45.29 \vline & 35.89 \\
            
            \cline{1-10}
            
            \multirow{5}{*}{BL}& $\checkmark$ & $\checkmark$ & $\circ$ & single & $\circ$ & $\circ$ & 59.49 \vline & 45.35 \vline & 34.71\\
 
            % & $\checkmark$ & $\checkmark$ & $\circ$ & $\checkmark$ & $\circ$ & $\circ$ & 77.95 \vline & 67.16 \vline & 17.43 \\
            
            & $\checkmark$ & $\checkmark$ & $\checkmark$ & bilateral & $\circ$  & $\circ$ & 59.84 \vline & 45.78 \vline & 33.19 \\
            
            &$\checkmark$ & $\checkmark$ & $\checkmark$ & $\checkmark$ & $\checkmark$ & $\circ$ & 60.08 \vline & 46.16 \vline & 33.05 \\
            
            & $\checkmark$ & $\checkmark$ & $\checkmark$ & $\checkmark$ & $\checkmark$ & step 1 & 60.38 \vline & 46.29 \vline & 32.66 \\
            
            & $\checkmark$ & $\checkmark$ & $\checkmark$ & $\checkmark$ & $\checkmark$ & step 1+2 & \textbf{60.71}\vline &\textbf{46.69}\vline & \textbf{31.91}\\
            \hline
        \end{tabular}
			
\end{table}
Baseline (BL) is a vanilla MT \cite{tarvainen2017mean}, with the remaining experimental settings consistent with those of the comparative experiments. In the second row of Table \ref{tab6}, the input to the student network is adjusted to the LF images, and the CT network will be trained on the EF images. At this stage, the pseudo-labels generated by the CT are used to supervise the outputs of the $\text{L}_s$ through single $\mathcal{L}_{fcs}$. The results show notable improvements, highlighting the effectiveness of the self-learning capability of the CT and the importance of learning specialized knowledge by $\text{L}_s$. In the third row, we introduce $\text{H}_s$, and the supervision by $\mathcal{L}_{fcs}$ becomes bidirectional. This operation further improves segmentation performance because $\text{H}_s$ successfully extracts features from HF and feeds this specialized expertise back to the CT. Next, we introduce $\mathcal{L}_{ccs}$ between $\text{L}_s$ and $\text{H}_s$, forming CFC-MT. Compared to using only $\mathcal{L}_{fcs}$, segmentation results are further improved, demonstrating the importance of knowledge exchange between the two SS networks.

In the last two rows of Table \ref{tab6}, we introduced UCF-Mix into training to validate its effectiveness. In the second-to-last row, we applied only the first step of UCF-Mix, while in the last row, we used both steps, forming the full CFC-Net. It can be seen that both results show improvements compared to the stage without UCF-Mix. Additionally, using the two-stage UCF-Mix leads to further improvement over the single-stage, fully demonstrating the effectiveness of UCF-Mix and the necessity of the bidirectional mix mechanism.

\section{CONCLUSION AND FUTURE WORK}
In this paper, we first introduce FMRC-2025, a expert-annotated CBCT dataset for SSL FMRC segmentation. Secondly, we propose a SSL network called CFC-Net. Extensive experiments confirm that its segmentation performance surpasses previous SOTA SSL networks. Furthermore, we evaluate CFC-Net on three public available dental datasets, demonstrating its strong robustness and generalizability across the dental segmentation tasks. Finally, we conduct ablation studies to validate the effectiveness of each component within CFC-Net.

In the future, our work will proceed in two directions: First, we plan to further improve the FMRC datasets by increasing their size and progressing toward full-mouth RC segmentation. Second, we will continue exploring the application of deep learning in RC segmentation tasks, focusing on enhancing segmentation accuracy and optimizing model architectures.

\section*{Acknowledgments}
This work is partially supported by the National Natural Science Foundation (62272248), the Natural Science Foundation of Tianjin (23JCZDJC01010).

\bibliographystyle{./elsarticle-harv.bst}
\bibliography{./ref.bib}

\end{document}